\documentclass[letterpaper, 10pt, conference]{ieeeconf}

\IEEEoverridecommandlockouts
\usepackage[utf8]{inputenc}

\usepackage{xspace}
\usepackage{comment}
\usepackage{graphicx}
\usepackage{xcolor}
\definecolor{gg}{RGB}{0, 155, 85}
\definecolor{primarycolor}{RGB}{33,49,77}   
\definecolor{angrycolor}{RGB}{210,73,42}    

\usepackage{todonotes}

\usepackage{amsmath}
\usepackage{amssymb}
\usepackage{amsthm}
\theoremstyle{definition}

\usepackage{tikz}
\usetikzlibrary{automata,positioning,angles,quotes}
\usepackage{tikz-cd}

\newtheoremstyle{main}
                {1em}                                                
                {1em}                                                
                {\normalfont}                                        
                {0pt}                                                
                {\scshape}                                           
                {\\*}                                                
                {2pt}                                                
                {\thmname{#1}\thmnumber{ #2}: \thmnote{\itshape #3}} 
                
\theoremstyle{main}

\makeatletter
\let\NAT@parse\undefined
\makeatother
\usepackage[pdfa,colorlinks,bookmarksopen,bookmarksnumbered,allcolors=gg]{hyperref}
\usepackage{cite}

\usepackage[algoruled,vlined,linesnumbered]{algorithm2e}

\SetAlgoSkip{SkipBeforeAndAfter}

\SetCommentSty{mycommfont}
\SetKw{Continue}{continue}
\usepackage{booktabs}
\usepackage{multirow}
\usepackage{makecell}

\usepackage[font=footnotesize]{caption}
\usepackage[font=footnotesize]{subcaption}
\usepackage[export]{adjustbox}

\usepackage[
activate   = {true},
protrusion = true,
expansion  = true,
kerning    = true,
spacing    = true,
tracking   = false,
auto       = true,
selected   = true,
factor     = 1000,
stretch    = 15,
shrink     = 15,
]{microtype}

\newcommand{\R}[1]{{\ensuremath{\mathbb{R}^{#1}}}\xspace}
\newcommand{\SEthree}{\ensuremath{SE(3)}\xspace}
\newcommand{\SEtwo}{\ensuremath{SE(2)}\xspace}
\newcommand{\SOthree}{\ensuremath{SO(3)}\xspace}

\newcommand{\dof}{\textsc{dof}\xspace}

\newcommand{\C}{{\ensuremath{\mathcal{C}}}\xspace}
\newcommand{\Cs}{\ensuremath{\mathcal{C}}-space\xspace}

\newcommand{\T}{{\ensuremath{\mathcal{T}}}\xspace}
\newcommand{\Ts}{\ensuremath{\mathcal{T}}-space\xspace}

\newcommand{\F}{\ensuremath{\mathcal{F}}\xspace}

\newcommand{\abbr}[1]{\textsc{\MakeLowercase{#1}}\xspace}

\newcommand{\prm}{\abbr{PRM}}
\newcommand{\smm}{\abbr{SMM}}

\newcommand{\grr}{\abbr{grr}}
\newcommand{\rgrr}{\abbr{random-grr}}
\newcommand{\egrr}{\abbr{expansion-grr}}

\title{\LARGE \bf
Expansion-GRR: Efficient Generation of Smooth Global 

Redundancy Resolution Roadmaps
}
\author{Zhuoyun Zhong, Zhi Li, and Constantinos Chamzas
  \thanks{%
   All authors are affiliated with the Department of Robotics Engineering, Worcester Polytechnic Institute (WPI), Worcester, MA 01609, USA {\tt\small \{zzhong3, zli11, cchamzas\} @ wpi.edu}.
  }
}

\usepackage{draftwatermark}
\SetWatermarkText{Accepted for publication in IEEE/RSJ International Conference on Intelligent Robots and Systems (IROS), October 2024}

\SetWatermarkColor[gray]{0.5}
\SetWatermarkFontSize{0.42cm}
\SetWatermarkAngle{0}
\SetWatermarkHorCenter{0.5\paperwidth}
\SetWatermarkVerCenter{0.5in}

\begin{document}

\maketitle
\thispagestyle{empty}
\pagestyle{empty}

\begin{abstract}
Global redundancy resolution (\grr) roadmap is a novel concept in robotics that facilitates the mapping from task space paths to configuration space paths in a legible, predictable, and repeatable way.
Such roadmaps could find widespread utility in applications such as safe teleoperation, consistent path planning, and motion primitives generation.
However, previous methods to compute \grr roadmaps often necessitate a lengthy computation time and produce non-smooth paths, limiting their practical efficacy.
To address this challenge, we introduce a novel method \egrr that leverages efficient configuration space projections and enables a rapid generation of smooth roadmaps that satisfy the task constraints.
Additionally, we propose a simple multi-seed strategy that further enhances the final quality.
We conducted experiments in simulation with a 5-link planar manipulator and a Kinova arm.
We were able to generate the \grr roadmaps up to 2 orders of magnitude faster while achieving higher smoothness. 
We also demonstrate the utility of the \grr roadmaps in teleoperation tasks where our method outperformed prior methods and reactive IK solvers in terms of success rate and solution quality.

\end{abstract}

\section{Introduction}
\label{sec:introduction}

\begin{figure}
    \centering
    \vspace{0.5em}
    
    \includegraphics[width = \linewidth]{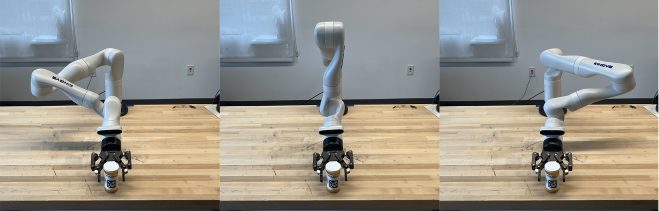}
    \caption{The Kinova arm is tasked with reaching a 6-\dof pose to pick up an object. For this task, the 7-\dof arm is kinematically redundant and can reach the same object position with multiple configurations.}
    \label{fig:global1}
    
    \vspace{-1em}
\end{figure}

A robot is considered kinematically redundant for a specific task if it has more degrees of freedom (\dof) than those strictly required by the task \cite{chiaverini_redundant_2015}. As demonstrated in \autoref{fig:global1}, the 7-\dof Kinova robotic arm is asked to reach a target object with a specific end-effector pose. Given its redundancy, the robot can accomplish this task through multiple or potentially infinite configurations. 
In general, the challenge of redundancy resolution is determining which among the many configurations to select as the most appropriate for the task at hand.

\begin{figure}
    \centering

    \begin{subfigure}[bt]{0.75\linewidth}
        \centering
        \includegraphics[width = \linewidth]{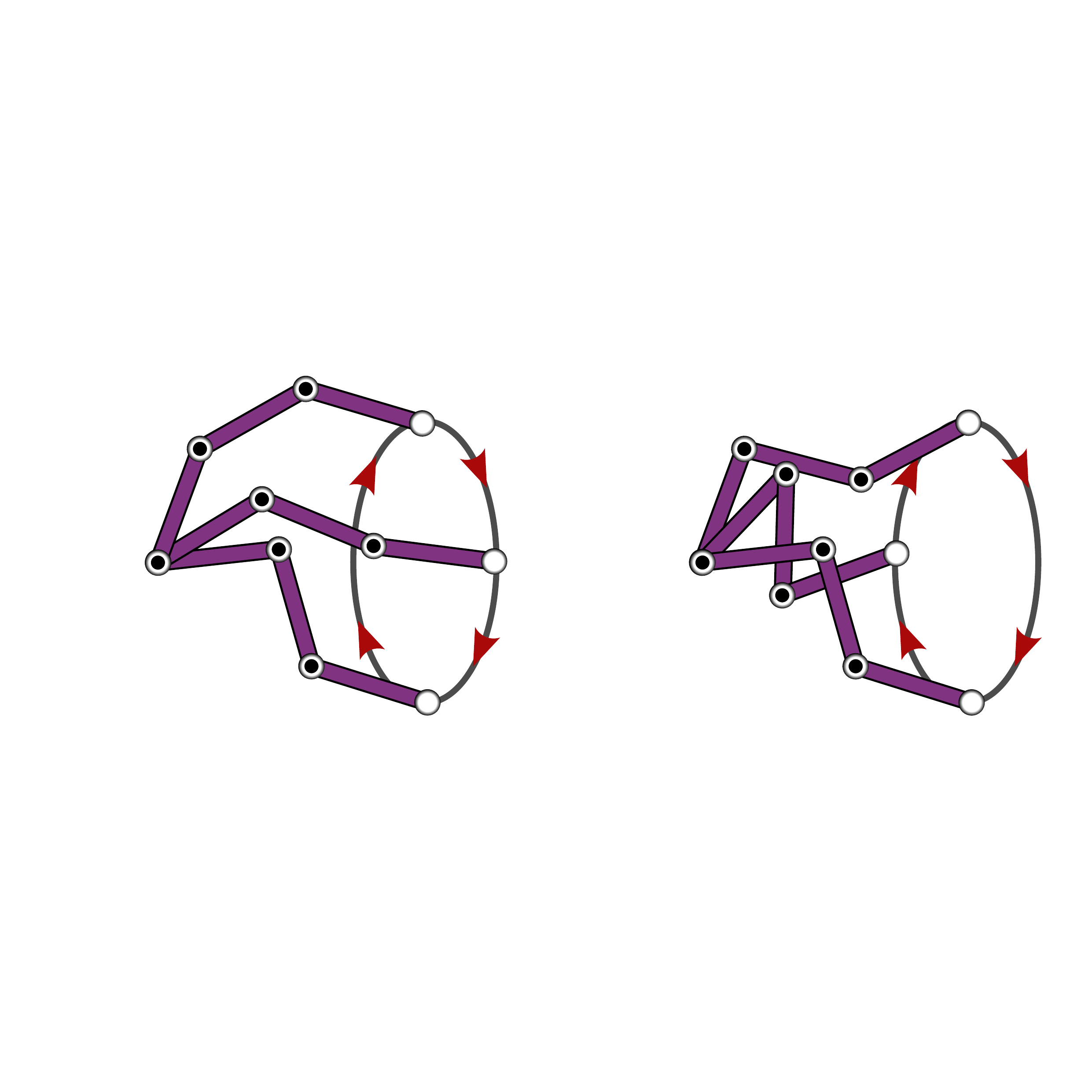}
        \caption{Non-global Resolution}
        \label{fig:global2-1}
    \end{subfigure}
    
    \begin{subfigure}[ht]{0.75\linewidth}
        \centering
        \includegraphics[width = \linewidth]{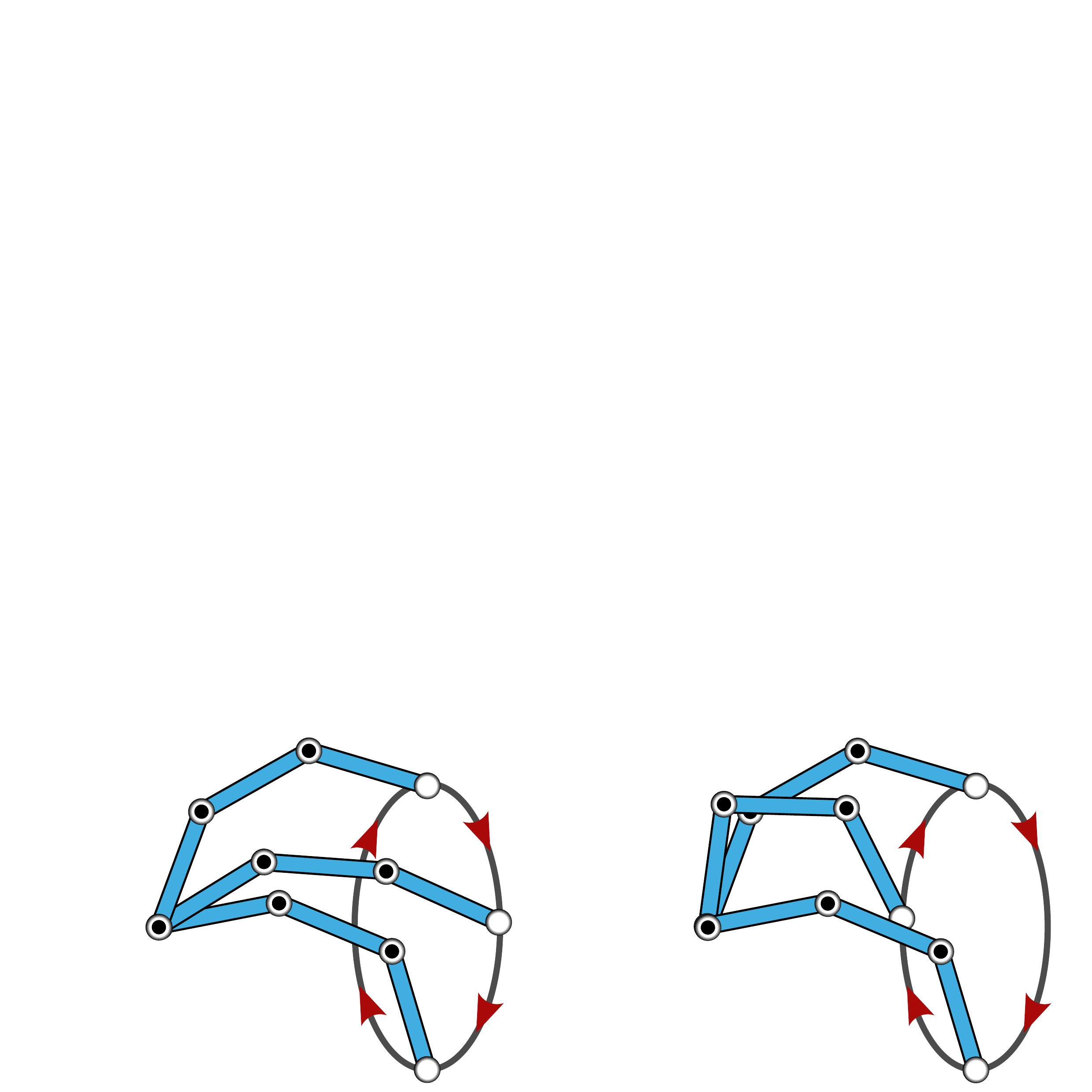}
        \caption{Global Resolution}
        \label{fig:global2-2}
    \end{subfigure}

    \caption{(a) After a cyclic path, non-global resolution can lead to different ending configurations from the same starting point. (b) Global resolution results in consistent paths, where the robot always returns to the original configuration.}
    \label{fig:global2}

    \vspace{-2em}
\end{figure}

In this work, we focus on the problem of global redundancy resolution (\grr) for end-effector paths \cite{hauser_global_2018}. 
This resolution enables a useful consistency property such that after any cyclic paths, it will always lead the robot back to the same configuration. For example, as shown in \autoref{fig:global2}, upon executing a closed ellipse path, the global method will always lead the robot back to the original configuration while the non-global method may take the robot to another one. This feature ensures the robot's motion is always repeatable and more predictable during execution.

\grr roadmaps can potentially serve various purposes in robotics, including streamlining repeatable tasks and motion primitives generation. Moreover, a primary application lies in teleoperation interfaces for humans. A common strategy of teleoperation is to allow operators to issue motion commands directly to the end effector space, while the teleoperation system translates these commands into actual joint motion. A \grr roadmap can aid such a system more effectively than reactive Inverse Kinematics (IK) solvers, as it can effectively recover from singularities and entrapment of local minima, while exhibiting repeatable and predictable motions. 

However, the computation of such a roadmap poses significant computational challenges. Previously proposed methods often required several hours to days to complete the process~\cite{hauser_global_2018} and often yield \grr roadmaps with lower smoothness. 
To tackle this issue, we propose a novel method \egrr that uses an expansion strategy to rapidly connect adjacent nodes through configuration space projections. To ensure continuity, we consider multiple nearest neighbors during expansion. Also, we propose a multi-seed strategy to further enhance roadmap quality.

Moreover, we introduce a straightforward yet effective approach to demonstrate the practical utility of this roadmap in teleoperation tasks. This method remains applicable even in scenarios where the \grr roadmap lacks full continuity or where users direct the robot to perform actions that might lead to illegal maneuvers, such as encountering a self-collision. Through our experiments in simulation, we illustrate that the proposed method not only generates higher-quality roadmaps more efficiently but also outperforms advanced IK solvers in teleoperation tasks, involving robots with up to 7 \dof. 

Specifically, the contributions of this work include
1) the development of a new expansion algorithm for efficiently computing a smooth \grr roadmap,
2) the proposal of a multi-seed strategy resulting in higher quality \grr roadmaps,
and 3) the establishment of a teleoperation pipeline utilizing the \grr roadmap that surpasses prior methods.
The implementation of the proposed algorithm\footnote{\url{https://github.com/elpis-lab/Expansion-GRR}} is provided as open source. 

\section{Related Work}
Redundancy resolution has historically been a core concept in robotics \cite{khatib_unified_1987}. Its applications span a wide spectrum, including avoidance of singularities \cite{chiaverini_singularity-robust_1997} and obstacles \cite{maciejewski_obstacle_1985} as well as facilitating teleoperation tasks \cite{brooks_assistance_2023}. Traditionally, redundancy is resolved at the velocity level \cite{klein_review_1983
} or the torque level \cite{khatib_unified_1987} through the utilization of projection operators. 
However, our work focuses on redundancy resolution at the position level \cite{albu-schaffer_redundancy_2023, hauser_global_2018} where we compute a direct mapping from task-space point to robot configurations.

A commonly used strategy for addressing redundancy at the local position level is IK solvers.
Examples include simple Jacobian-based IK solvers such as the Newton-Rapson method
or more recent IK-solvers, such as Relaxed-IK \cite{rakita_relaxedik_2018} and Ranged IK \cite{wang_rangedik_2023}, which incorporate more complicated constraints including self-collisions and singularity avoidance.
These IK solvers exhibit high-speed performance, often processing at several iterations per second. However, they may get trapped in singularity or local minima during teleoperation and generate suboptimal solutions due to their lack of foresight and local problem-solving approaches.

Another category of methods that resolve redundancy at the position level while often avoiding the local minima is single-query planning methods \cite{orthey2024-review-sampling}. Several planning methods that also incorporate task-space constraints \cite{berenson_task_2011} have been proposed, with a commonly used constrained sampling-based motion planning framework \cite{kingston_sampling-based_2018}. Nonetheless, the produced motions are often not repeatable as shown in \autoref{fig:global2-1}. To address this challenge, \cite{oriolo_motion_2005, oriolo_repeatable_2017} focus on producing cyclic repeatable paths that start and end at the same configuration.
However, these methods are tailored towards producing a single repeatable path, while in this work, we aim to resolve all possible task-satisfying paths simultaneously.

Roadmap-based methods, e.g., probabilistic roadmaps (\prm)\cite{kavraki_probabilistic_1996}, compute an approximation of the connectivity of the configuration space through random sampling. Given a start and goal configuration, the roadmap can be queried with graph search, and retrieve a collision-free path.
Our work also computes a roadmap but is tailored towards producing paths that additionally satisfy task constraints and the paths are smooth both in task space and configuration space. The work that is most similar to ours is \cite{hauser_global_2018}, which formulated the global redundancy resolution problem, and proposed an algorithm to compute a \grr roadmap. In this work, we propose a new algorithm that generates the \grr roadmaps more efficiently and produces smoother paths. We also demonstrate how it can be used for real-time motion planning as in teleoperation.

Teleoperation is still a critical control mode of robotics, especially in safety-critical applications like surgery. 
There are many different teleoperating modes and interfaces depending on the application \cite{niemeyer_telerobotics_2016}. 
A common teleoperation interface is for humans to give commands in their natural 6-\dof task-space,
which then has to be retargeted, to a kinematically different (and often redundant) robot \cite{boguslavskii_shared_2023}.
To compute the retargeting, a common strategy is to use reactive IK-solvers \cite{
rakita_relaxedik_2018, wang_rangedik_2023}
and convert cartesian commands directly to the robot's configuration space. In this work, we showcase the utility of the computed \grr roadmap in teleoperation tasks, outperforming reactive-IK solvers, in terms of success rate and solution quality.

\section{Notations and Problem Statement}
\label{sec:problem}

\subsection{Definitions And Notations}

The general redundancy resolution problem aims to find a map $\mathcal{M}$ from the robot's Task Space (\Ts) to its Configuration Space (\Cs). \Ts represents the domain in which the robot performs its tasks,  typically  $R^3$ or $\SEthree$, while \Cs contains all potential configurations available to the robot. We will denote the elements of \Ts as
$p \in \T$ and the elements of \Cs as $q \in \C$.
Redundancy indicates that the dimension of \Cs is higher than that of \Ts, i.e., $dim(\C)~>~dim(\T)$.  

\subsubsection{Local Redundancy Resolution}
A map $\mathcal{M}$ is a local redundancy resolution if it is a transformation from $p \in \T$ to $q \in \C$ such that $\mathtt{FK}(q) = p$, where $\mathtt{FK}$ is the forward kinematics function. For example, IK solvers are local resolutions at the point-wise level, and single-query planning methods are local resolutions at the path level.

\subsubsection{Global Redundancy Resolution}
A map $\mathcal{M}$ is defined as global redundancy resolution if it satisfies the $\mathtt{FK}(q)=p$ constraint and additionally is an \emph{injective}, \emph{continuous} and \emph{smooth}\footnote{Technically, smoothness implies continuity, but since we measure them separately, we mention them both here.} function $\mathcal{F}:~\T~\to~\C$. The injectivity property ensures that each point in \Ts is uniquely associated with a single configuration in \Cs, maintaining the consistency of the result. Continuity and smoothness guarantee that a small change in \Ts does not lead to discontinuous or large changes in \Cs, resulting in continuous and smooth robotic motions.

\subsection{Approximate Global Redundancy Resolution}
\begin{figure}
    \vspace{0.5em}
    
    \begin{subfigure}[ht]{0.4\linewidth}
        \includegraphics[height=0.14\textheight,center]{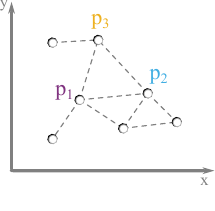}
        \caption{\Ts Roadmap $G_p$}
        \label{fig:roadmap1}
    \end{subfigure}%
    ~
    \begin{subfigure}[ht]{0.55\linewidth}
        \includegraphics[height=0.14\textheight,center]{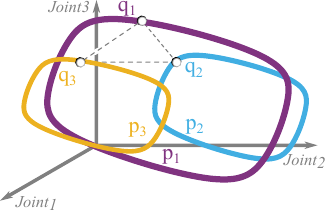}
        \caption{\Cs Roadmap $G_q$}
        \label{fig:roadmap2}
    \end{subfigure}

    \caption{An illustration of the roadmaps $G_p$ and  $G_q$. (a) The points $p_1, p_2$, and $p_3$ are neighboring samples in \Ts, in this case, the end-effector position in \R2. (b) The solid lines indicate \smm of \Ts points. A roadmap $G_q$ must select a single configuration $q_i$ from each manifold, and additionally, ensure that the configurations, $q_1, q_2$ and $q_3$ of adjacent points, are "close enough" to pertain continuity and smooth transition in \Cs.}
   \label{fig:roadmap}

   \vspace{-1em}
\end{figure}

Given that finding an analytical form for \F is intractable, the proposed method focuses instead on computing an approximation~\cite{hauser_global_2018}.
An illustrative example is shown in \autoref{fig:roadmap} for the same 3-link planar manipulator of \autoref{fig:global2} that operates in \R2.

We first approximate \Ts with a discrete roadmap graph $G_p=(V_p, E_p)$. For the 3-link manipulator example,  $G_p$ is shown in \autoref{fig:roadmap1}, with the \Ts points $p_1, p_2, p_3 \in V_p$ as nodes and corresponding edges shown as dashed lines.
In \autoref{fig:roadmap2}, the solid lines indicate the set of configurations $q$ that satisfy the constraints $\mathtt{FK}(q) = p$ induced by $p_1, p_2$, and $p_3$, also known as the self-motion manifolds (\smm) \cite{burdick_inverse_1989}.
To compute an approximation for \F, we need to build the corresponding \Cs roadmap $G_q$ that satisfies the \grr properties:  
\begin{enumerate}
    \item \textbf{Injectivity:} For each \Ts vertex $p_i \in V_p$, there is only a single corresponding \Cs vertex $q_i \in V_q$, i.e,  choosing a single configuration $q$ from each self-motion manifold (as shown in \autoref{fig:roadmap2}). 

    \item \textbf{Continuity:}  For each \Ts edge $(p_i, p_j) \in E_p$, the configurations of the matching \Cs edge $(q_i, q_j) \in E_q$ satisfy the continuity constraint,  formally defined in \autoref{sec:method-continuity} and shown in \autoref{fig:continuous}. 
    
    \item \textbf{Smoothness:} Minimize the length ratio of \Cs edges $(p_i, p_j) \in E_p$ to \Ts edges $(q_i, q_j) \in E_q$.
\end{enumerate}

Concretely, the problem we are concerned with is: Given a graph $G_p$, compute a graph $G_q$ that maximally satisfies 1), 2), and 3) while minimizing total computational time.

\section{Methodology}
\label{sec:methodology}
\SetAlgorithmName{Algorithm}{Alg.}

The main idea of the proposed method is to efficiently build the roadmap $G_q$ by expanding from the selected initial configuration seeds. 
Again, for illustrative purposes, we will employ a 3-link planar manipulator operating in \R2 task space (similar to that depicted in~\autoref{fig:global2})

\subsection{Continuity Constraint}
\label{sec:method-continuity}

To find a configuration $q$ on the self-motion manifold that satisfies the constraint imposed by a point $p$, i.e., $\mathtt{FK}(q) = p$, a common method employed is projection in \Cs~\cite{kingston_exploring_2019, albu-schaffer_redundancy_2023}. Starting with a given initial guess $q_{guess}$, an optimizer iteratively moves it closer to the desired self-motion manifold until a configuration $q$ within tolerance is found. In this paper, as our projection operator,  we use the Newton-Raphson IK, which is a Jacobian pseudo-inverse projection. Also, to ensure the returned configuration is valid a self-collision check is performed. We denote this function as $\mathtt{Projection}(p, q_{quess})$.

\begin{figure}
    \vspace{0.5em}

    \begin{subfigure}[ht]{0.4\linewidth}
        \includegraphics[height=0.14\textheight,center]{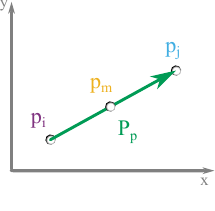}
        \caption{Path in \Ts}
        \label{fig:continuous1}
    \end{subfigure}%
    ~
    \begin{subfigure}[ht]{0.55\linewidth}
        \includegraphics[height=0.14\textheight,center]{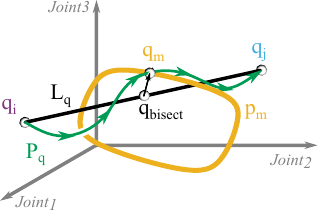}
        \caption{Path in \Cs}
        \label{fig:continuous2}
    \end{subfigure}

    \caption{An illustration of the continuity constraint. (a) The robot's end-effector follows a straight path $P_p$ from $p_i$ to its adjacent point $p_j$ in \Ts (green). (b) The robot's corresponding configuration path $P_q$ from $q_i$ to $q_j$ in \Cs (green) can be computed by projecting the intermediate configurations to their corresponding \smm in a bisection manner. Ideally, $P_q$ should stay "close" to the straight line $L_q$ delineated by $q_i$ and $q_j$ (black).}
    \label{fig:continuous}

    \vspace{-1em}
\end{figure}

Continuity ensures that a small variation in \Ts does not lead to a discontinuous motion in \Cs. To achieve this, we enforce a motion constraint similar to \cite{taylor_planning_1979}. An illustration with the 3-link planar manipulator is provided in~\autoref{fig:continuous}.
Let us assume we want to move the end effector from point $p_i$ to its adjacent point $p_j$ in a straight line $P_p$ in \Ts. The motion constraint ensures that the corresponding configuration path $P_q$ will not deviate significantly from the straight line $L_q$ delineated by $q_i$ and $q_j$ in \Cs.

\autoref{fig:continuous} illustrates visually how to perform this check. We project the bisected configuration $q_{bisect}$ to the self-motion manifold of the middle point $p_m$ and acquire the intermediate configuration $q_m$. To pass the continuity check, the deviation between $q_m$ and $q_{bisect}$, should be within a threshold. As shown in~\autoref{alg:continuous} this process is repeated recursively to ensure the continuity check passes up to an $\epsilon$ resolution.

\begin{algorithm}
\caption{$\mathtt{IsContinuous}(p_i, p_j, q_i, q_j)$}
\label{alg:continuous}
\SetNoFillComment

\tcc{Check threshold}
\If{$\mathtt{D_q}(q_i, q_j) < \epsilon$}{
    \Return $true$\;
}

\tcc{Bisect in \Ts and \Cs}
$p_m \gets \mathtt{BisectP}(p_i, p_j)$\;
$q_{bisect} \gets \mathtt{BisectQ}(q_i, q_j)$\;
$q_m \gets \mathtt{Projection}(p_m, q_{bisect})$\;

\tcc{Check deviation}
\If{$\mathtt{Max}(\mathtt{D_q}(q_i, q_m), \mathtt{D_q}(q_m, q_j)) > c \cdot \mathtt{D_q}(q_i, q_j)$}{
    \Return $false$\;
}

\tcc{Run recursively}
\If{$\mathtt{IsContinuous}(p_i, p_m, q_i, q_m)$ \\ \normalfont{and} $\mathtt{IsContinuous}(p_m, p_j, q_m, q_j)$}{
    \Return $true$\;
}
\Return $false$\;

\end{algorithm}

In \autoref{alg:continuous}, $\mathtt{BisectP}$ finds the center of two \Ts points, and $\mathtt{BisectQ}$ finds the center of two \Cs configurations. The parameter $c > 0.5$, regulates the maximum deviation. The value $\epsilon$ is the minimal threshold to stop the visibility checking process. In this paper, we choose $c$ and $\epsilon$ based on the dimension of the \Cs, i.e., $c = 0.5 \sqrt{(dim(\C))}$, and $\epsilon = 0.05 \sqrt{(dim(\C))}$. These values are chosen as reasonable default values that scale with the dimension of the \Cs.

\subsection{Projection From Multiple Neighbors}
\label{sec:method-projection}
\begin{figure}
    \vspace{0.5em}

    \begin{subfigure}[ht]{0.4\linewidth}
        \includegraphics[height=0.14\textheight,center]{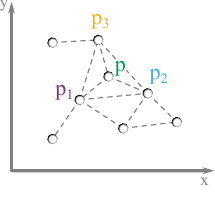}
        \caption{\Ts Roadmap $G_p$}
        \label{fig:average1}
    \end{subfigure}%
    ~
    \begin{subfigure}[ht]{0.55\linewidth}
        \includegraphics[height=0.14\textheight,center]{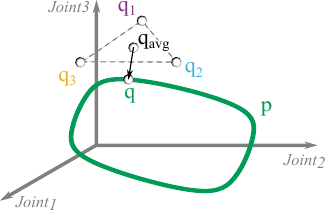}
        \caption{\Cs Roadmap $G_q$}
        \label{fig:average2}
    \end{subfigure}
    
    \caption{  An illustration of projecting from multiple neighbors. (a) The points $p_1, p_2$, and $p_3$ are neighboring points of $p$ in \Ts. (b) The solid lines indicate the self-motion manifold of $p$. A weighted average $q_{avg}$ is computed with neighbors' corresponding configurations $q_1$, $q_2$, and $q_3$ in \Cs. It is then projected onto the self-motion manifold to find $q$.}
    \vspace{-2em}

    \label{fig:average}
\end{figure}

Now we describe the proposed method for adding new configurations in an existing $G_q$ roadmap, also illustrated with an example in \autoref{fig:average}.

One strategy is to choose a configuration from a \Ts neighbor of $p$ e.g., $q_3$, and project it to get $q$, i.e, $q = \mathtt{Projection}(p, q_3)$.  
However, since point $p$ has multiple neighbors, projecting from a single one will create a bias towards this neighbor. For example,  $\mathtt{Projection}(p, q_3)$ could result in a configuration that is far from $q_1$ and $q_2$, resulting in discontinuities. Instead, we adopt a more general strategy of computing a weighted average configuration $q_{avg}$ in \Cs from all neighboring configurations $q_1$, $q_2$, and $q_3$ as shown in \autoref{fig:average}. By aggregating neighboring configurations, the resulting $q$ derived from $\mathtt{Projection}(p, q_{avg})$ has a higher chance of being similar to all its neighbors. This similarity feature aids in improving overall smoothness and continuity.

We now explain this process more formally, with the pseudo-code in  \autoref{alg:project}.  
First, we denote a function that returns the $k$ nearest neighbors $neighbors$ of point $p$ in the \Ts graph $G_p$, as $\mathtt{NearestNeighbors}(p, k, G_p)$. 
We also define the \Ts distance metric $\mathtt{D_p}$ adapted from \cite{kuffner_effective_2004} as:
\begin{equation}
\begin{gathered} 
    \mathtt{D_p} (p_i, p_j) = w_t \cdot \mathtt{D_{\R3}} (t_i, t_j) + w_o \cdot \mathtt{D_{\SOthree}} (o_i, o_j), \text{where} \\
    \mathtt{D_{\R3}} (t_i, t_j) = \left \| t_i - t_j \right \| \\
    \mathtt{D_{\SOthree}} (o_i, o_j) = 1 - \left| o_i \cdot o_j \right| \\
\end{gathered} 
\end{equation}
and $t$, $o$ denote the translation and orientation (represented by a quaternion) component of a \Ts point. $\left \| t_i - t_j \right \|$ is the standard Euclidean norm and $\left| o_i \cdot o_j \right|$ is the absolute value of the inner product, while $w_t$ and $w_o$ are chosen weights to scale the distances. In this paper, we set $w_t = 1$ and $w_o = 0.3$. If $\T = \R3$, we set $w_o = 0$.

After locating the $neighbors$ of a point $p$, we collect a list of configurations $qs$ from them. Their corresponding weights $ws$ are determined based on their \Ts distances $ds$ and scale towards the closer neighbors. Lastly, we compute the weighted average configuration $q_{avg}$ and project it to find $q$.

\begin{figure}[ht]
    \centering
    \vspace{0.5em}

    \begin{subfigure}[ht]{0.475\linewidth}
        \centering
        \includegraphics[width = \linewidth]{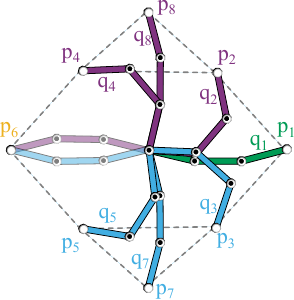}
        \caption{Starting with One Seed}
        \label{fig:expansion1}
    \end{subfigure}%
    ~
    \begin{subfigure}[ht]{0.475\linewidth}
        \centering
        \includegraphics[width = \linewidth]{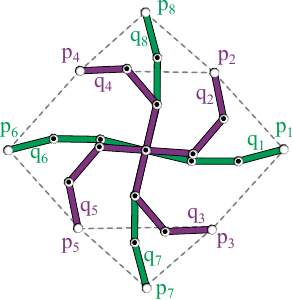}
        \caption{Starting with Multiple Seeds}
        \label{fig:expansion2}
    \end{subfigure}

    \caption{
    (a) From an initial seed $q1$ (green), $q_2$, $q_8$ to $q_4$ are sequentially generated (purple), and $q_3$, $q_7$ to $q_5$ are also sequentially generated (blue). For $q_6$, however, expanding from either side will lead to a configuration that is not continuous to the other.
    (b) Starting from seeds $q_1, q_7, q_8$ and $q_6$ (green), deriving from an initial cyclic path, we could achieve a fully connected \Ts roadmap with all ``elbow-down'' configurations (purple).
    }

\end{figure}

\begin{algorithm}
\caption{$\mathtt{ProjectNeighbors}(p, G_p, G_q$)}
\label{alg:project}
\SetNoFillComment

\tcc{Collect $qs$ and $ds$ from $neighbors$}
$neighbors \gets \mathtt{NearestNeighbors}(p_i, k, G_p)$\;
$qs \gets$ \normalfont{Empty configuration list}\;
$ds \gets$ \normalfont{Empty distance list}\;
\ForEach{$j \in neighbors$}{
    \If{$q_j$ \normalfont{is in} $G_q$}{
        $qs.append(q_j)$\;
        $ds.append(D_p(p, p_j))$\;
    }
}

\tcc{Compute weights}
$ws \gets$ \normalfont{Weight list with the same size as $qs$}\;
\ForEach{$j \in neighbors$}{
    $w_i \gets (\mathrm{Max}(ds) / d_i)^2$
}
$\mathtt{Normalize}(ws)$\;

\tcc{Compute and project from $q_{avg}$}
$q_{avg} \gets \mathtt{WeightedAverage}(qs, ws)$\;
\Return $\mathtt{Projection}(p, q_{avg})$\;

\end{algorithm}
\vspace{-1ex}

\subsection{Seeding Strategy}
\label{sec:seeding}
The techniques discussed in the previous sections can be used to expand the roadmap $G_q$ by incrementally adding new nodes.
However, the choice of the initial configuration(s) that the roadmap is seeded with, also has a significant impact on the final result.   
A simple seeding strategy would be to initialize $G_q$ with a random configuration and expand from there.
Yet, this simple seeding method is prone to create discontinuities in \Cs.
\autoref{fig:expansion1} illustrates an example with the simple 3-link manipulator. Starting from an initial seed $q_1$ such that $\mathtt{FK}(q_1) = p_1$, one set of generated configurations are ``elbow-down'' configurations (blue), while the other set is ``elbow up'' (purple). This results in a discontinuity at the $p_6$ task point, since no configuration (either purple or blue) will satisfy the continuity constraint of both sides.

Ideally, the seeds should be situated within \Cs to minimize the probability of encountering discontinuities.
To this aim, we propose an alternative multi-seeding strategy, as depicted in \autoref{fig:expansion2}.
Given that the final roadmap must encompass all potential cyclic paths, we propose to initialize the roadmap with configurations from a continuous cyclic path, shown as the green configurations in \autoref{fig:expansion2}.
While there exist algorithmic approaches to generating cyclic paths \cite{oriolo_motion_2005, oriolo_repeatable_2017}, we opted to manually create one per robot, since only a single cyclic path is required for our purposes.

\subsection{Global Expansion}
In this section, we describe how we can use the tools mentioned in the previous sections to build  
the \Cs roadmap $G_q = (V_q, E_q)$. We start by discretizing \Ts and generating the \Ts roadmap $G_p = (V_p, E_p)$. Although several choices exist, in this paper we use grid discretization. 
Given the selected initial configurations $qs_{init}$ and roadmap $G_p$, ~\autoref{alg:global} describes, how we incrementally expand the roadmap in \Cs by projection from nearest neighbors in a Breadth First Search (BFS) manner. 

First, we initialize the \Cs roadmap $G_q$, a queue $Q$, and the visited set $V$ with $qs_{init}$ (\autoref{global:queue}-\autoref{global:set}). Similar to BFS, it iteratively dequeues an index $i$ from $Q$ and explores its $k$ nearest neighbors in the \Ts (\autoref{global:BFSstart}-\autoref{global:BFSend}).
For each point $p_i$ that does not have a corresponding configuration, it uses \autoref{alg:project} to project the weighted average configuration and find the solution $q_i$ for the point $p_i$ (\autoref{global}). Finally, it runs $\mathtt{IsContinuous}$(\autoref{alg:continuous}) checks with all neighbors to verify continuity, and the results will be used to update $G_q$ (\autoref{global:Continuous}-\autoref{global:Add}).


\begin{algorithm}
\caption{$\mathtt{GlobalExpansion}(G_p, qs_{init})$}
\label{alg:global}
\SetNoFillComment

\tcc{Initialize}
Initialize $\Cs$ roadmap $G_q$ with $qs_{init}$\;   \nllabel{global:queue}
$Q \gets$ A queue with each $i \in qs_{init}$\;     
$V \gets$ A set with each $i \in qs_{init}$\;       \nllabel{global:set}
\While{$Q \neq \emptyset$}{                         \nllabel{global:BFSstart}
    
    \tcc{Expand in BFS order}
    $i \gets Q.dequeue()$\;
    \ForEach{$j \in $ $\mathtt{NearestNeighbors}(p_i, k, G_p)$}{
        \If{$j \notin V$}{
            $V.add(j)$\;
            $Q.enqueue(j)$\;
        }
    }                                               

    \If{$q_i$ \normalfont{is in} $G_q$}{            
        \Continue\;                                 \nllabel{global:BFSend}
    }

    \tcc{Project from neighbors}
    $q_i \gets \mathtt{ProjectNeighbors}(p_i, G_p, G_q)$\; \nllabel{global}

    \tcc{Continuity check and update}
    Run $\mathtt{IsContinuous}$ check with each neighbor\;     \nllabel{global:Continuous}

    Add $q_i$ and valid edges into $G_q$\;          \nllabel{global:Add}
}

\end{algorithm}

\vspace{-1ex}

\subsection{Utilizing The Roadmap}
Now, we introduce some potential use cases for the computed \grr roadmap.

\subsubsection{IK Solver}
\label{sec:use-ik}

To use the \grr map as an IK solver, we need to generalize it to the continuous space for all $p \in \T$. To this aim, the process described in \autoref{alg:project} can still be employed to find the corresponding configuration $q$ for any given point $p$. However, the functionality of $\mathtt{NearestNeighbor}$ will be different when discontinuous edges are present in the local neighboring subgraph of $G_q$. The function should only consider neighbors in the largest connected component of the subgraph, ignoring any disconnected neighbors. Another way to view this process is to use the \grr roadmap to get a good $q_{guess}$, and then use an off-the-shelf IK-solver to project it to the given task-point constraint.

\subsubsection{\Ts Path Planning}
\label{sec:use-planning}

To find a path from a starting configuration to a goal configuration, we can directly plan a path in the configuration roadmap $G_q$.
However, if it is desired to have the end-effector follow the edges of the \Ts roadmap $G_p$ closely, we can instead search and plan a path $P_p$ in roadmap $G_p$. We then interpolate the points along $P_p$ and for each interpolated waypoint $p$, we can run $\mathtt{ProjectNeighbors}$ to find its corresponding configuration $q$ and build the \Cs path $P_q$. This ensures the end effector will closely follow the edges of $G_p$. Also, since the interpolation happens along the edges, the configuration path $P_q$ will always be feasible and continuous because the edges have passed the $\mathtt{IsContinuous}$ check.

\subsubsection{Teleoperation}
\label{subsubsec:teleop}

Given a real-time Cartesian command from a human operator, we can produce the corresponding configuration as described in \autoref{sec:use-ik}. However, since the human input will not necessarily follow the edges of the roadmap, the produced configuration $q$ is not guaranteed to be feasible. It can get into self-collision or a discontinuous region if one exists. In this case, we can leverage the roadmaps and run planning as in \autoref{sec:use-planning} to detour from these regions.

We design the teleoperation pipeline as follows. Given the current configuration $q_{c}$ and a new human input $p_{t}$ as the target, we compute the next configuration $q_{t}$ by running $\mathtt{ProjectNeighbors}$. We verify the feasibility of going from $q_{c}$ to $q_{t}$ by testing it with $\mathtt{IsContinuous}$. We execute it if feasible. Otherwise, instead of going towards the human input $p_{t}$, we change the target from $p_{t}$ to $p_{t}$'s valid nearest neighbor $p_{n}$ in $G_p$. Such a change ensures the robot still follows human commands closely without getting into an infeasible zone. Once the human input $p_{t}$ becomes feasible again, we leverage planning as described in \autoref{sec:use-planning} to plan a path from $q_{c}$ to the new valid target $q_{t}$ and continue tracking human input.

\section{Experiments}

\begin{table*}[ht]
\centering
\vspace{0.5em}
\caption{Roadmap Quality}
\label{tab:roadmap}

\begin{tabular}{@{}l|ccc|lccc@{}}
\toprule
\multicolumn{1}{c|}{Robot} & Dim($\C$) & Dim($\T$) & Vertices & \multicolumn{1}{c}{Method} & Connectivity (\%) & Smoothness & Building Time (min) \\ \midrule

\multirow{2}{*}{\begin{tabular}[c]{@{}l@{}}5-link Manipulator\\ Position\end{tabular}} & \multirow{2}{*}{5} & \multirow{2}{*}{2} & \multirow{2}{*}{1,013} & \rgrr & \textbf{100.00} & 16.853 & 2.023 \\
 &  &  &  & \egrr & \textbf{100.00} & \textbf{5.675} & \textbf{0.095} \\ \midrule
 
\multirow{2}{*}{\begin{tabular}[c]{@{}l@{}}5-link Manipulator\\ Position \& Fixed Rotation\end{tabular}} & \multirow{2}{*}{5} & \multirow{2}{*}{3} & \multirow{2}{*}{1,013} & \rgrr & \textbf{100.00} & 24.528 & 1.862 \\
 &  &  &  & \egrr & \textbf{100.00} & \textbf{8.992} & \textbf{0.150} \\ \midrule

\multirow{2}{*}{\begin{tabular}[c]{@{}l@{}}Kinova\\ Position\end{tabular}} & \multirow{2}{*}{7} & \multirow{2}{*}{3} & \multirow{2}{*}{3,299} & \rgrr & \textbf{100.00} & 9.223 & 22.971 \\
 &  &  &  & \egrr & \textbf{100.00} & \textbf{2.563} & \textbf{0.201} \\ \midrule
 
\multirow{2}{*}{\begin{tabular}[c]{@{}l@{}}Kinova\\ Position \& Fixed Rotation\end{tabular}} & \multirow{2}{*}{7} & \multirow{2}{*}{6} & \multirow{2}{*}{3,299} & \rgrr & \textbf{100.00} & 11.205 & 72.013 \\
 &  &  &  & \egrr & \textbf{100.00} & \textbf{4.299} & \textbf{0.196} \\ \midrule

\end{tabular}

\vspace{-2em}
\end{table*}


We evaluate the performance of our method in two aspects. First, we assess the quality of the \grr roadmap relative to the criteria set in \autoref{sec:problem}. Second, we emulated teleoperation tasks and compared the results obtained from the \grr roadmaps and different IK-solvers. All the experiments were run on a machine with Intel i5-10400 and 16 GB memory.

\subsection{Roadmap Quality Metrics}

To measure the quality of the resolution roadmap $G_q$, two metrics are used:

\subsubsection{Roadmap Connectivity}
A fully connected \Cs roadmap $G_q$ may not exist or may be difficult to find, resulting in a discontinuous graph. We quantify this concept of  connectivity with the metric:

\begin{equation}
    C(G_q) = \frac{N_{E_q}}{N_{E_p}}
\end{equation}
where $N_{E_q}$ is the size of the \Cs edge set $E_q$, and $N_{E_p}$ is the size of the \Ts edge set $E_p$.
Thus, a fully connected roadmap $U(G_q) = 1$ refers to a \Cs roadmap that has as many edges as the \Ts roadmap.

\subsubsection{Roadmap Smoothness}
Another metric that evaluates the roadmap quality is the smoothness, approximated by the averaged ratio of \Cs edge distance to \Ts edge distance:

\begin{equation}
    S(G_q) = \frac{1}{N_{E_q}}\sum_{(i, j)\in E_q} \frac{\mathtt{D_q}(q_i, q_j)}{\mathtt{D_p}(p_i, p_j)}
\end{equation}
where $N_{E_q}$ is the size of the \Cs edge set $E_q$. $\mathtt{D_q}$ and $\mathtt{D_p}$ are functions to compute the distance in \Cs and \Ts.

\subsection{Roadmap Quality Experiment}

We compare the roadmap connectivity, smoothness, and building time with the \grr method proposed in~\cite{hauser_global_2018}. Instead of using multi-seeding and projections, \cite{hauser_global_2018} builds the \grr map by first randomly sampling numerous \Cs configurations at each \Ts point. It then connects all the adjacent pairs that pass the continuity test. Eventually, a constraint satisfaction problem (\textsc{CSP}) solver is used to select a single configuration for each point.
We refer to this method as \rgrr. In all experiments, we used 100 random samples per \Ts point. 

We tested in simulation two robots, a 5-link planar manipulator (5-\dof), and a  Kinova Gen3 arm (7-\dof) with a Robotiq-85 gripper (shown in \autoref{fig:global1}. The 5-link planar manipulator joints are continuous (have no limits) and self-intersections are allowed. The tested \Ts for the planar manipulator are $\T = \R2$ and $\T = \SEtwo$, and the latter incorporates a fixed rotation where the end-effector is always facing right.
The Kinova robot was tested with $\T = \R3$ and $\T = \SEthree$, with the latter incorporating a fixed rotation where the end-effector is consistently facing downwards.


The results are given in~\autoref{tab:roadmap}. Both \rgrr and \egrr can fully resolve the global redundancy. However, \egrr outperforms \rgrr by building a smoother roadmap up to 2 orders of magnitude faster.  We attribute the improved smoothness to the proposed projections which minimize the distances between neighboring configurations. Also, by utilizing multi-seeding and projection, we can avoid the time-consuming steps caused by extensive sampling in random-\grr.

\subsection{Teleoperation Experiment}

To evaluate the performance of our method in teleoperation, we designed four geometric path tracing tasks, which provide commands to robots in \Ts:
\begin{itemize}
    \item Random Line: A line defined by two random points.
    
    \item Self-crossing Line: A line as in Random Line, but it passes through the robot's base.
    
    \item Random Circle: A planar circle defined by a random center point, up vector, and radius.
    
    \item Partially Reachable Circle: A planar circle as in Random Circle, but part of it goes beyond the reachable space.
\end{itemize}

For each task, 100 paths are randomly generated. Intermediate waypoints are sent sequentially to the solvers at 50 Hz over 4 seconds, i.e. 200 waypoints per path to follow, simulating a human input action.

To use \egrr and \rgrr for teleoperation we use the teleoperation pipeline we proposed in \autoref{subsubsec:teleop}, and compare against Newton-Raphson IK and Relaxed IK~\cite{rakita_relaxedik_2018}. The weights of the Relaxed IK terms were chosen to minimize deviation from input as suggested in~\cite{rakita_relaxedik_2018}.
We measure the following three metrics:

\subsubsection{Deviation From Input}
The deviation from the produced path to the input path in \Ts is measured by finding the pairs of matched waypoints with dynamic time warping and then computing the average distance between these pairs. 

\subsubsection{Path Smoothness}
We define path smoothness similar to roadmap smoothness, as the average distance ratio of the produced \Cs path to that of the produced \Ts path.

\subsubsection{Success Rate}
The success is determined by whether the produced \Ts path follows the input path to the goal. We marked it as a failure if the produced path gets stuck due to local minima or self-collision.

For a fair comparison, we calculate deviation and smoothness only for the problems that all methods succeed.

We ran the tasks with the Kinova arm in $\T = \R3$ with fixed rotation and the aggregated results are summarized in~\autoref{tab:trajectory}. 
The IK solvers often got stuck in local minima at line-related tasks and encountered singularities in circle-related tasks. \rgrr had the lowest success rate, as it tended to produce discontinuous paths from the non-smooth map.
The proposed method \egrr was able to achieve $100\%$ success rate in all types of tasks. 
Regarding deviation from input, \egrr performed as well as Newton-IK in tasks Random-Line and Random-Circle, but in the more challenging tasks Self-Crossing Line and Partially Reachable Circle, our method exhibited higher deviation due to the need for more planning.
For smoothness, the IK-solvers achieved the best results for each \Ts path. This result is expected as our method prioritizes global consistency and continuity over smoothness. Nevertheless, \egrr still produces smoother paths compared to \rgrr.

\begin{table}[h]
\centering
\vspace{0.5em}
\caption{Teleoperation Experiment}
\label{tab:trajectory}

\begin{tabular}{c|c|ccc}
\hline
Type & Methods & \begin{tabular}[c]{@{}c@{}}Deviation\\ From Input\end{tabular} & \begin{tabular}[c]{@{}c@{}}Path\\ Smoothness\end{tabular} & \begin{tabular}[c]{@{}c@{}}Success\\ Rate \%\end{tabular} \\ \hline
\multirow{4}{*}{\begin{tabular}[c]{@{}c@{}}Random\\ Line\end{tabular}} & Newton IK & \textbf{0.011} & \textbf{4.395} & 90 \\
 & Relaxed IK & 0.023 & 4.569 & 78 \\
 & \rgrr & 1.984 & 6.432 & 49 \\
 & \egrr & \textbf{0.011} & 5.071 & \textbf{100} \\ \hline
\multirow{4}{*}{\begin{tabular}[c]{@{}c@{}}Self-\\ crossing\\ Line\end{tabular}} & Newton IK & 0.479 & 7.882 & 30 \\
 & Relaxed IK & \textbf{0.184} & \textbf{4.236} & 54 \\
 & \rgrr & 2.547 & 5.760 & 34 \\
 & \egrr & 0.461 & 5.481 & \textbf{100}\\ \hline
\multirow{4}{*}{\begin{tabular}[c]{@{}c@{}}Random\\ Circle\end{tabular}} & Newton IK & \textbf{0.022} & 4.284 & 98 \\
 & Relaxed IK & 0.023 & \textbf{3.986} & 98 \\
 & \rgrr & 0.458 & 6.974 & 77 \\
 & \egrr & \textbf{0.022} & 4.664 & \textbf{100}\\ \hline
\multirow{4}{*}{\begin{tabular}[c]{@{}c@{}}Partially\\ Reach-\\ able\\ Circle\end{tabular}} & Newton IK & \textbf{0.085} & 6.138 & 99 \\
 & Relaxed IK & 0.102 & \textbf{3.997} & 95 \\
 & \rgrr & 0.652 & 6.428 & 78 \\
 & \egrr & 0.166 & 5.200 &\textbf{100}\\ \hline
\end{tabular}

\vspace{-4ex}
\end{table}

\section{Discussion}
\label{sec:discussion}
In this paper, we presented a new method to generate smooth \grr roadmaps in an efficient manner. We also demonstrated these properties compared to the prior art and illustrated how they can be used for teleoperation tasks.

One of the limitations of our work is that it uses all of the available redundancy of the robot to satisfy the \Ts constraints, making it challenging to meet other secondary objectives.
In the future we would like to address this problem by building more flexible roadmaps similar to \cite{yao_edm_2024, albu-schaffer_redundancy_2023}. The selection of initial seeds is another interesting problem to investigate. One potential avenue is to generate cyclic paths using \cite{oriolo_motion_2005, oriolo_repeatable_2017}. Although our method achieved full connectivity in the tested examples, this might not always be possible e.g., when obstacles are present. Nonetheless, we expect the roadmap to still be usable as in the example with the infeasible query of the self-crossing lines.

Finally, we would like to investigate applications of \grr roadmaps beyond teleoperation such as motion planning, dimensional reduction, and generating motion primitives.


\bibliographystyle{IEEEtran}
\bibliography{references_manual, references}

\end{document}